\def\year{2022}\relax
\documentclass[letterpaper]{article} 
\usepackage{aaai22}  
\usepackage{times}  
\usepackage{helvet}  
\usepackage{courier}  
\usepackage[hyphens]{url}  
\usepackage{graphicx} 
\usepackage{threeparttable}
\usepackage{natbib}  
\usepackage{caption} 
\DeclareCaptionStyle{ruled}{labelfont=normalfont,labelsep=colon,strut=off} 
\usepackage{algorithm}
\usepackage{algorithmic}
\usepackage{amsmath}
\usepackage{multirow}
\usepackage{makecell}

\usepackage{xcolor}

\usepackage{newfloat}
\usepackage{listings}
\floatstyle{ruled}
\newfloat{listing}{tb}{lst}{}
\floatname{listing}{Listing}
\usepackage[colorlinks,citecolor=blue,urlcolor=blue,bookmarks=false,hypertexnames=true]{hyperref}

\title{Visual-Semantic Transformer for Scene Text Recognition}
\author{
    Xin Tang\equalcontrib,
    Yongquan Lai\equalcontrib,
    Ying Liu,
    Yuanyuan Fu,
    Rui Fang
}
\affiliations{
    Visual Computing Group, Ping An Property $\&$ Casualty Insurance Company, Shenzhen, China \\

    tangxin051@pingan.com.cn,tangxint@gmail.com
}

\usepackage{bibentry}

\begin{document}

\maketitle

\begin{abstract}
Modeling semantic information is helpful for scene text recognition. In this work, we propose to model semantic and visual information jointly with a Visual-Semantic Transformer (VST). The VST first explicitly extracts primary semantic information from visual feature maps with a transformer module and a primary visual-semantic alignment module. The semantic information is then joined with the visual feature maps (viewed as a sequence) to form a pseudo multi-domain sequence combining visual and semantic information, which is subsequently fed into an transformer-based interaction module to enable learning of  interactions between visual and semantic features. In this way, the visual features can be enhanced by the semantic information and vice versus. The enhanced version of visual features are further decoded by a secondary visual-semantic alignment module which shares weights with the primary one. Finally, the decoded visual features and the enhanced semantic features are jointly processed by the third transformer module obtaining the final text prediction. Experiments on seven public benchmarks including regular/ irregular text recognition datasets verifies the effectiveness  our proposed model, reaching state of the art on four of the seven benchmarks.
\end{abstract}

\section{Introduction}

Scene text recognition (STR) is the task of recognizing text from images taken in complex scenes such as street view. It is an inherently difficult task due to the variance of shape, color, scale and appearance of the embedded text. Clutter background, large perspective distortion,  lighting condition, degraded image quality due to motion/out-of-focus blur also impose severe challenges to the successfully solving the task,  resulting severe miss-predictions. 

Despite its difficulty, STR has many real-world applications ranging from self-driving cars \cite{AD}, street image understanding to applications such as instant translation and intelligent text reading in smart-phones \cite{TT}.   For decades, STR has been an active research direction \cite{TR,SAR,ASTER, AI0,AI1}, attracting many efforts in designing new models and creating new datasets in order to solve the problem.

A complete approach to recognizing text from scene images usually involves text detection and text recognition. In this paper, we assume that text detection is done and only focus on the recognition part. That is, we assume the input to our model is a cropped image with regular or irregular characters lying in it.  A large portion of the  previous work and many open-source datasets \cite{IC13, IC15} follow from this assumption. In this work, we will refer to STR as just recognizing text from cropped images.

The approaches to solving STR problem can be roughly divided into two categories: linguistic-based and linguistic-free. Linguistic-based methods refer to those which incorporate vocabulary (dictionary) or lexicon (parts of words), while linguistic-free methods use only the images themselves without relying on explicit language modeling, whether from internal or external,  pretrained or from-scratch.

The \textbf{motivation} of this work is multi-fold. 
Firstly, to deal with the cases when visual information alone is inadequate, semantic or linguistic features has been introduced in various effects, among which \cite{qiao2020seed} propose a semantic
enhanced encoder-decoder framework to recognize
low-quality scene texts. A semantic module is designed to directly produce semantic features  that are consistent with the word embedding learned from a pretrained language model.  Inspired by their efforts in exploiting semantic features, we also explicitly model semantic information .  But unlike their approach, we achieve semantic modeling without  relying on external language models but instead using an alignment module. 

On the other hand,  our work is also partially inspired from wav2vec \cite{baevski2020wav2vec} in audio community. The wav2vec model first extract primary audio features from waveform using 1D convolution.  The features are processed by a succeeding transformer module, resulting in secondary contextualized audio features. The secondary audio features interact with the primary features by predicting them back at the next few time steps. Similarly, we extract  semantic and visual features at different stages, making them interact with each other so that the overall recognition performance is improved.
\begin{figure*}[!h]
	\includegraphics[width=18cm]{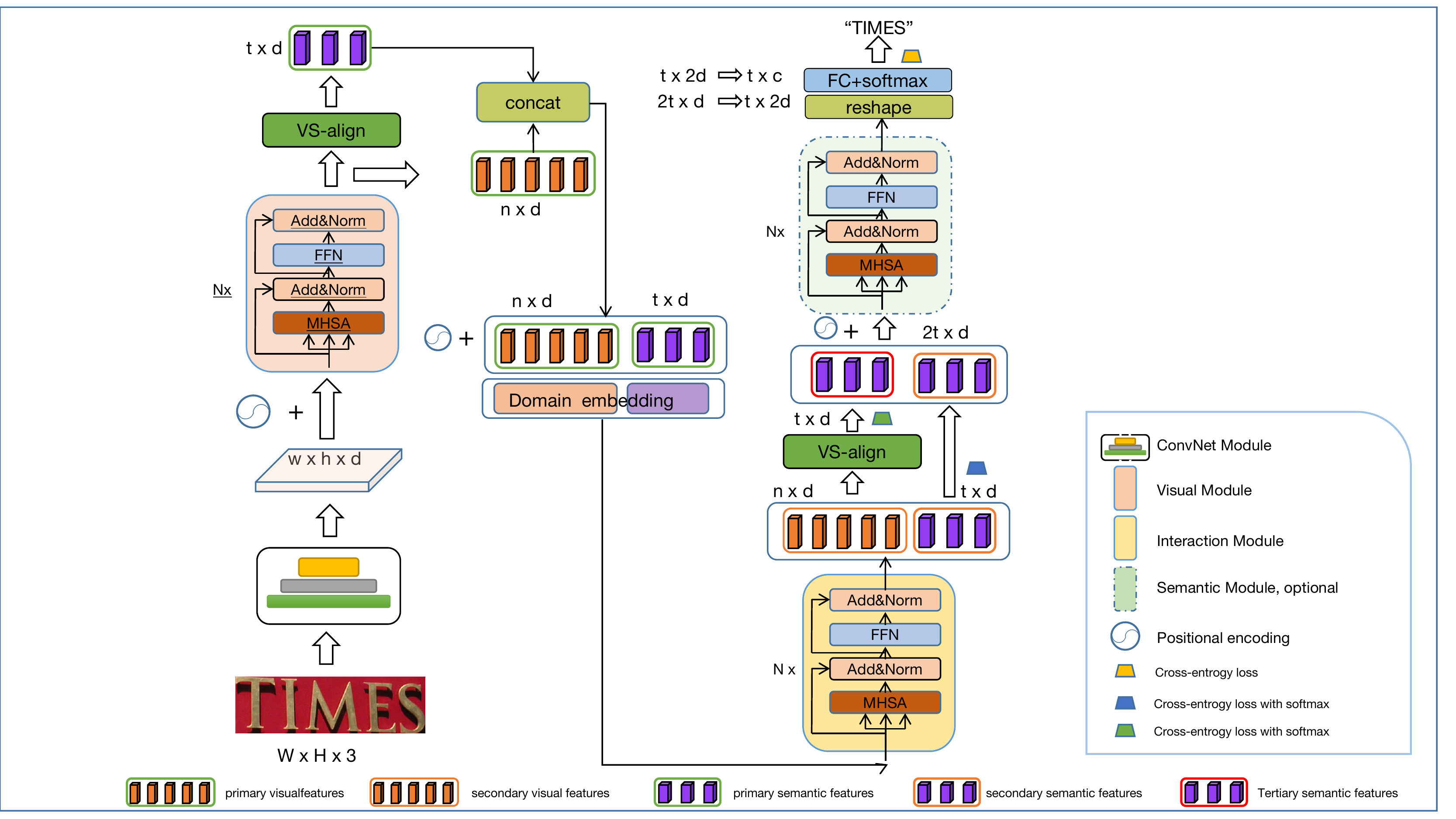}
	\caption{The architecture of Visual-Semantic Transformer (VST).  The VST consists of several key modules, namely ConvNet (C), Visual module (V), Interaction module (I), Semantic module (S) and two weight-sharing Visual-Semantic Alignment (A, or vs align) modules.  Best view in color.}
	\label{arch}
\end{figure*}

Apart from the aforementioned work, we have also noted that many efforts have been devoted to modeling visual-semantic relation.  We argue that extracting semantic features from appearance and then enforce interaction with visual features is one key to successfully solving STR. To this end, we propose a novel transformer model for scene text recognition. Our model is a unified end-to-end text recognizer which converts image patches into text in parallel (i.e., non-autoregressively), without using complex decoder such as CTC \cite{graves2006connectionist}.  The model, which we coin \textit{Visual-Semantic Transformer (VST)}, is able to  learn  semantic features from appearance and combine them back with visual features to enable visual-semantic interaction using  multi-head self-attentions (MHSA).

 The overall architecture of VST is shown in Fig. \ref{arch}.  The VST first explicitly extracts primary semantic information from visual feature maps with a transformer module and a primary visual-semantic alignment module (vs-align). The semantic information is then joined with the visual feature maps (viewed as a sequence) to form a pseudo multi-domain sequence combining visual and semantic information, which is subsequently fed into the second transformer module to enable learning of  interactions between visual and semantic features. In this way, the visual features can be enhanced by the semantic information and vice versus. The visual features are further decoded with a secondary vs-align module which shares weights with the primary one. Finally, the decoded visual features and the semantic features can be jointly processed by the third transform module and the final softmax layer to obtain the resulting text. Overall, our approach is end-to-end and conceptually simple. Experiments on public benchmarks on regular/ irregular text recognition tasks demonstrate the effectiveness of our proposed model. The main \textbf{contributions} of this paper are as follows, 
\begin{itemize}
\item We propose a novel visual-semantic transformer to effectively solve STR problem, surpassing or on par with state-of-the-art in most STR datasets.
 \item We design weight-sharing visual-semantic alignment modules to explicitly enforce the learning of semantic information without external language models.
\item  We introduce an interaction module that allows visual and semantic information to globally interact with and enhance from each other.
\end{itemize}

In this work, \textbf{semantic information} refers to the information that connects visual appearance and the underlying linguistic information. In other words, it is the information extracted from visual features which are very closely related to the text represented by the scene text image. Semantic information is distinguished from the term language or linguistic information, because the later usually refers to the information extracted directly from text solely. Semantic feature extraction is not language modeling or word embedding because  features are not computed directed from real character sequence, but rather learned from image appearance and forced to be consistent with language. The semantic feature is ready to convert into text characters using a simple linear probe. Under this definition, our model can be categorized as linguistic-free, as we do not require an external language model.  Instead, we explicitly model semantic information which can be seen as pseudo-linguistic information. 

In the following sections, we will briefly review the related work and then introduce our VST model in details, followed by extensive experiments and conclusion.

\subsection{Related Work}
Text recognition has been an active research area for decades.  See \cite{AI1,chen2021text} for comprehensive reviews. Due to page limitation, we can only list a portion of recent work here. 

Recently, 
\cite{nguyen2021dictionary}  incorporate a dictionary in training and inference stage to  help selecting the most compatible outcome for STR. \cite{feng2021semantic} introduce character center segmentation branch to extract semantic
features which encode the category and position of characters for improving video text detection performance. \cite{patel2016dynamic} propose to
generate contextualized lexicons for scene images with only
visual information. \cite{sabir2018visual}  use language model to build the
semantic correlation between scene and text in order to re-rank the
recognition results. \cite{zheng2019deep} also propose to use pretrained language
models for correcting predictions. Very recently, \cite{wang2021two} propose to learn the linguistic rules in the visual space by randomly mask out some characters from the image and predict them back.

Similar to speech recognition, scene text recognition can be treated as a sequence-to-sequence (seq2seq) mapping problem \cite{DEEP2014,DEEP2015,SEED,RobustScanner}.
\cite{SAR} combine convolution and LSTM as an encoder, then use another LSTM as decoder to predict text attentively, quite similar to the \textit{show, attend and tell}  work on image captioning \cite{xu2015show}. 
  ASTER \cite{ASTER} takes a two-stage approach to first rectify curved text images and then perform recognition using seq2seq model with attention. CRNN \cite{CRNN} adapts   CNN to obtain visual features, which are then  fed into LSTM module with CTC loss \cite{graves2006connectionist} for  text prediction. STAR-Net \cite{STAR-Net} uses spatial transformer \cite{STN} to tackle challenges brought by image distortion.  Attention can be added to the seq2seq models \cite{SAR, Bhunia2021JointVS} in a straightfoward way to alleviate  bottleneck effects brought by seq2seq models.  \cite{litman2020scatter}  propose stacked block architecture with intermediate supervision to train a deep BiLSTM encoder, while attention is used in decoding stage to exploit contextualized visual features. \cite{aberdam2021sequence}  propose seq2seq contrastive learning of visual representations which can be applied to text recognition. DAN \cite{DAN} decouples alignment from the decoding stage into the early conventional encoder network.
 \cite{wang2021implicit} propose an alignment module enabling text recognizer to recognize document level image.  \cite{TextSR}  use adversarial loss to handle low-resolution image.

 Transformers have also been successfully applied to STR.  ABINet \cite{RLH} enforces the bidirectional language-model (LM) to only learn linguistic rules by gradient-stopping in training. The decoding is in an iterative way allowing the predictions to be refined progressively. Their conv+transformer visual module and transformer-LM can be separately pretrained to improve performance.  HRGAT \cite{HRGAT} 
connects CNN feature maps to a transformer-based autoregressive decoder, where the cross-attention is 
guided by holistic representation obtained by average-pooling of 2d feature maps.

Perhaps more related,  SRN \cite{SRN} incorperates visual-to-semantic embedding block and cross-entropy loss to align with ground-truth text, but they use argmax embedding, which is different from our direct use of probability vector that enables smooth gradient flow in training. Our work is also innovative in many ways such as visual-semantic alignment, multiple-stage semantic processing and transformer-based visual-semantic interaction. Instead of using argmax, \cite{Bhunia2021JointVS}  use Gumbel-softmax \cite{jang2016categorical}  for extracting semantic information, which is then fed into succeeding transformer-based visual-semantic reasoning module. The decoding involves complex multiple-stage attentional LSTM that couples with feature pyramid networks. Our approach is not only conceptually much simpler and computationally more efficient, but also more effective in solving STR.

\section{Visual-Semantic transformer}

\begin{figure}[h]
	\includegraphics[width=0.49\textwidth, trim=0.0cm 1.0cm 1.5cm 0.0cm]{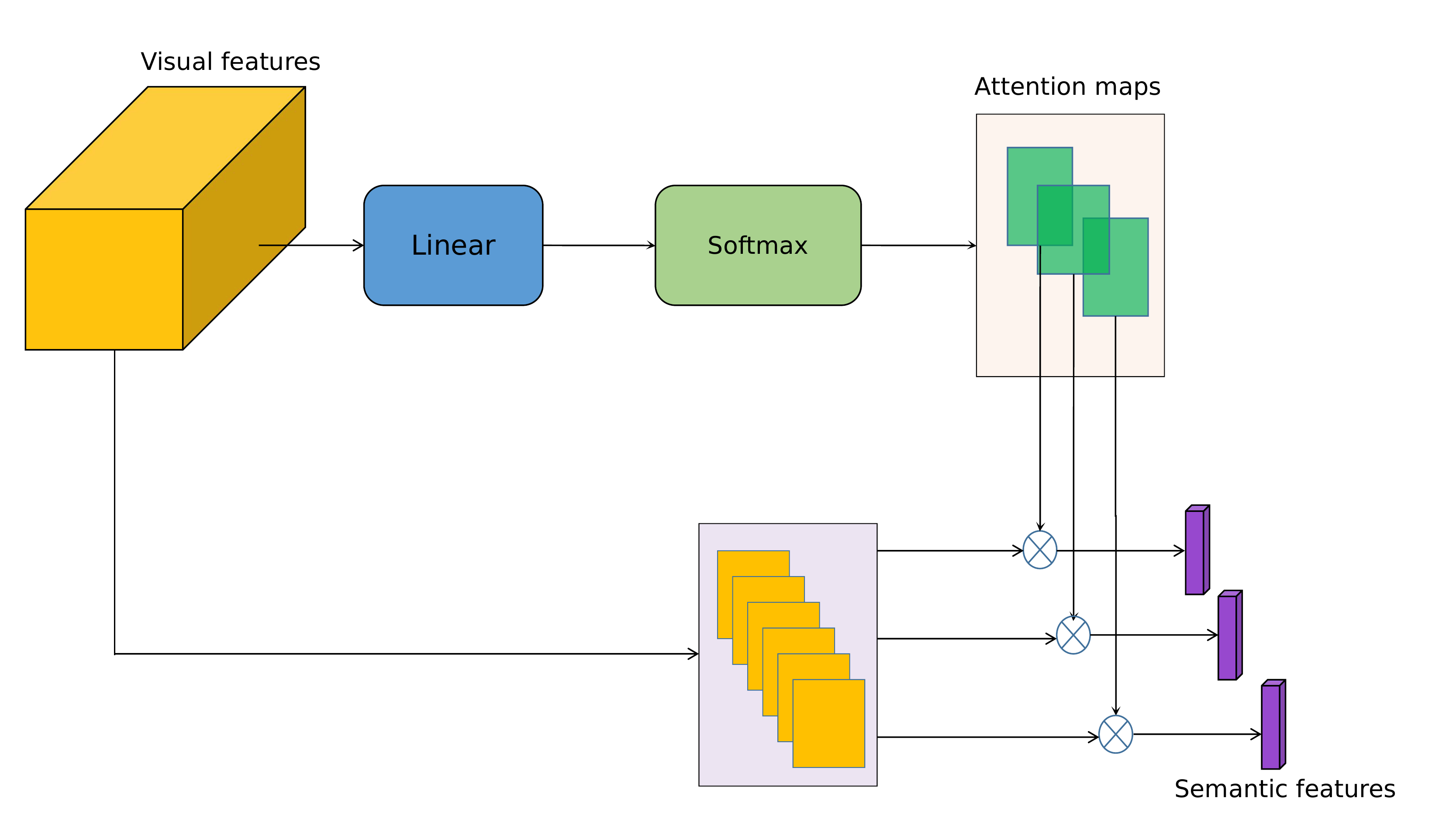}
	\caption{The architecture of vs-align module. Visual features (viewed as feature maps) are projected using a linear layer and further normalized using sofmtax operator, obtaining $t$  attention maps, each of which has the same spatial dimension as the origin feature map.  The $t$ attention maps and the original $d$ visual feature maps together will reduce (by multiplication) to $t$ semantic features in $R^d$.}
	\label{fig:vs_align}
\end{figure}
We will introduce the visual-semantic transformer (VST) in this section .
The VST consists of several key modules, namely ConvNet (C), Visual module (V), Interaction module (I), Semantic module (S) and two weight-sharing Visual-Semantic Alignment (A, or vs-align) module.  Module C can be any convnet, Module V, I, S are basically transfomer blocks, while vs-align is an attention-based alignment block. Since a large portion of our model consists of transformer blocks,  we name it \textit{visual-semantic transformer}, implying that it is a transformer that explicitly models visual and semantic information.

The overview  of the architecture is shown Fig. \ref{arch}. The convnet module extracts local visual features from cropped images. The  visual module transforms the local visual features into 
contextualized global visual features, which we call primary visual features. The first vs-align module converts the primary visual features into primary semantic features. The interaction module then transforms primary visual and primary semantic features  by allowing them to interact with each other through MHSA,  producing secondary visual and secondary semantic features respectively. The secondary visual features are fed into the second vs-align module (which shares weights with the first one), obtaining  tertiary (third) semantic features.  Secondary and tertiary semantic features together will be processed by the final semantic module to obtain the resulting text prediction.  

To enforce the learning of semantic information, the two vs-align modules \textbf{share weights}.  In other words, the improvement of  the second vs-align module is immediately helpful for improving the accuracy in aligning the primary semantic features, which in return will help improve the overall recognition accuracy. We found that this design is very useful for extracting primary semantic information, as seen in Fig. \ref{fig:atten_map}.

The VST has two variants, namely VST-F (full) or VST-B (basic), depending on whether the module S is used or not. In VST-F, there are three losses in training and the output of final softmax is used for text decoding. For VST-B, there are two losses  (blue and green in Fig. \ref{arch}) and the semantic stream of module I's output are passed to softmax layer, while the visual stream will go through additional vs-align module before softmax and CE loss. The decoding for VST-B can be evaluated separately or jointly (by probability voting).
\subsection{The ConvNet (C) Module}
 We use resnet-like architecture in extracting local features.
The input images are first resized to have the same height, with aspect ratio kept unchanged. During training and testing, replication-padding (padding using values from the image border) is used for batch processing. 

Assuming that input image is of  size $W \times H \times 3$, the convnet module will produce feature maps of size $w \times h\times d$. The features will be fed into 
the visual module.

\subsection{The Visual (V) Module}

Convnets are good at learning local features but hard to 
learn global correlation between features that locate far apart. Inspired from the success of transformers when applied to vision task, we insert a transformer module here  in order to enhance the feature maps by allowing them to  interact with each other. The visual modules takes the convolution feature maps as input, producing primary visual features.  The primary visual features together encode completely the appearance information of the input image.  They are then converted to primary semantic features by the vs-align module.

 The visual module consists of MHSA, layer-norm, and feed-forward network, as described in \cite{Attention}, but with minor modification that puts layer-norm before MHSA \cite{dosovitskiy2020image,wang2019learning}. Note that Fig. 1 does NOT reflect this modification. The feature maps are viewed/reshaped as a sequence of dimension $d$ and length $n=w \times h$. the module V, takes the feature  sequence as input and generate a sequence of the same dimension and length. The interaction between visual features themselves mainly takes place in the MHSA layers.

\subsection{The Visual-Semantic Alignment (A) Module}

Parallel attentions has been used to map visual features into semantics\cite{wang2021two,lyu20192d}. Compared with previous work, the vs-align module in this work is much simpler but still effective.  
 Note that we do not use the name semantic decoder here because the output of the module is not actually the character index,  but rather the probability distribution of the characters. In other words, the semantic feature encodes the probability distribution at each candidate location, which can be readily converted into character index with $argmax$ operation in decoding stage (not in training). This allows the gradients to flow smoothly inside the VST. 

The architecture of  vs-align is depicted in Fig. \ref{fig:vs_align}. As shown in the figure, the visual features (viewed as feature maps) are projected using a linear layer and normalize using softmax operator, obtaining $t$ attention maps each of which has the same spatial dimension as the origin feature map.
The alignment module can also be formulated mathematically as follows, 
\begin{equation}\label{eq:vsa}
S = \text{softmax}(QV^T)V
\end{equation}
where  $S \in R^{t\times d}$ is the semantic sequence, $V \in R^{n\times d}$ is the visual sequence and $Q \in R^{t\times d}$ is the trainable projection matrix.

Our VS-align module is able to learn the relation between the semantic and visual features. This is different from CTC \cite{graves2006connectionist}, which is good at aligning two sequences with beam search, but not capable of handling information along the height direction.

As shown in Fig. \ref{arch}, there are two vs-align modules, which share weights during training and inference.  This weight-sharing scheme is the key to successfully learning 1st semantic information. It enables the weights learned in aligning 2nd visual features to transfer to aligning 1st visual features, making the early  learning of 1st semantics possible. The 1st semantic in return enhances the 1st visual features through interaction module, further improving the training of the 2nd vs-align module. The  early learning of semantics provides more chances to correct the semantic features through interaction in later stages. If semantic had been learned in the last stage, we would have not change to correct it, unless incorporating additional lexicon or dictionary information
\cite{nguyen2021dictionary}

\begin{table*}
	\begin{tabular}{l|cccc|ccc}
		\hline
		\multirow{2}*{Method}&
		\multicolumn{4}{c|}{Regular test datasets}&
		\multicolumn{3}{c}{Irregular test datasets}\\
		&IIIT & SVT & IC03 & IC13&IC15&SVTP&CUTE  \\
		\hline
		AON~\cite{cheng2018aon} & 87.0&82.8&91.5&\_&68.2&73.0&76.8\\
		ASTER~\cite{ASTER}  &93.4&89.5&94.5&91.8&76.1&78.5&79.5 \\
		NRTR~\cite{NRTR} &86.5 &88.3  &95.4&94.7&\_&\_&\_\\
		SAR~\cite{SAR} &91.5 &84.5&&91.0&69.2&76.4&83.3\\
		DAN~\cite{DAN} &94.3 &89.2&95.0&93.9&74.5&80.0&84.4\\
		HRGAT~\cite{HRGAT} &94.7 &88.9&\_&93.2&79.5&80.9&85.4\\
		SRN ~\cite{SRN} &94.8 &91.5&\_&95.5&82.7&85.1&87.8\\
		SCATTER~\cite{SCATTER}  &93.7 &92.7&96.3&93.9&82.2&86.9&87.5\\
		GTC~\cite{Hu2020GTCGT}&95.5&92.9&95.2&94.3&82.5&86.2&92.3\\
		RobustScanner~\cite{RobustScanner}  &95.3 &88.1&\_&94.8&77.1&79.5&90.3\\
		~\cite{Bhunia2021JointVS} &95.2 &92.2&\_&95.5&84.0&85.7&89.7\\
		PREN~\cite{yan2021primitive}  &95.6 &94.0&95.8&96.4&83.0&87.6&91.7\\
		ABINet-SV~\cite{RLH}  &95.4 &93.2&\_&96.8&84.0&87.0&88.9\\
		ABINet-LV~\cite{RLH}  &96.2 &93.5&\_&\textbf{97.4}&\textbf{86.0}&\textbf{89.3}&89.2\\
		VST-B& \textbf{96.3} &\textbf{93.8} &96.4 &96.4 &85.4 &88.7 &\textbf{95.1} \\
		VST-F&96.1 &93.1&\textbf{97.1}&96.4&85.4&89.1&94.8\\
		\hline
	\end{tabular}
	\centering
	\caption{When compared with previous work, our approach achieves very competitive results. VST-B denotes the  C+V+A+I basic model and VST-F  the C+V+A+I+S full model.  '-' denotes  data not available or config not the same.}
	\label{tab:sota}
\end{table*}
\subsection{The Interaction (I) Module}

The interaction module plays the key role of mixing primary semantic features and visual features. The modules takes two streams , namely the primary semantic features and primary visual features as input,  producing the secondary visual and semantic features.  Fixed positional encoding are added into the visual stream. For the semantic stream,  learnable position embedding is used.  Other positional encoding schemes probably help as long as it breaks permutation invariant property of transformers.

The design of the interaction module follows from module V,  comprising layers of transformer blocks. The module enables the feature interaction between a) the visual stream and semantic 
 will interact with each other, and  b) the features themselves inside the same stream. 

The interpretation of module can be seen from eq. \ref{eq:inter}.  Following from the notations in \cite{Attention}, the interaction  in the MHSA layer is obvious by judging form the following equation, 
\begin{equation}\label{eq:inter}
S = \text{softmax}(\frac{[Q_s;Q_v][K_s;K_v]^T}{\sqrt{d_k}})[V_s;V_v],
\end{equation}
where  the subscripts s and v denote semantic and visual part respectively, ';' is column representation.

In this way, the primary semantic features will look for support from the spatial visual features.  The interaction is useful for enhancing semantic features because they  now have attentive access to all spatial location of the visual features. On the other hand, visual features are also enhanced not only because then can interact with all other spatial locations regardless of distance,  but they can also learn from semantic features, which we hypothesize is useful for dealing with appearance degradation. 

The semantic features can be seen as pseudo-linguistic features,  distinguished from visual domain. Hence we add domain embedding before feeding two combined streams into the interaction module.

One of the  outputs of interaction module is secondary visual sequence in $R^{n \times d}$, which will be converted to tertiary semantic features using vs-align module following by cross-entropy loss with softmax, against the ground-truth text. The other is  semantic features in $R^{n \times d}$, which will be directly compared against ground-truth text with the same activation ans loss type. 
Note that at inference time, the two loss branches are unneeded.

\begin{table}[t]
\begin{threeparttable}
	\resizebox{0.48\textwidth}{!}{
	\begin{tabular}{|l|cccc|ccc|}  
		\hline
		\multirow{2}*{Module}&
		\multicolumn{4}{c|}{Regular test datasets} &
		\multicolumn{3}{c|}{Irregular test datasets} \\
		& IIIT & SVT & IC03 & IC13 & IC15&SVTP&CUTE \\
		\hline
		CV&95.10 & 91.94  &95.72&95.52 &81.83&86.23&91.33   \\
		CVA&95.63 & 91.94  &96.41&95.55 &82.31&86.95&91.76   \\
		CVAI /$s_3$&96.40&93.51&96.54&96.35&85.09&88.84&94.79\\
		CVAI /$s_2$&96.26&\textbf{93.81}&96.30&96.35&85.14&88.68&95.13\\
		CVAI /p&\textbf{96.36} & \textbf{93.81} &96.42&\textbf{96.45} &85.36&88.68&\textbf{95.13}  \\
		CVAIS&96.06&93.50&\textbf{97.11}&96.40&\textbf{85.37}&\textbf{89.15}&94.79\\
		\hline
	\end{tabular}}
	\centering
	\caption{Recognition accuacy increases when extra module is added.  /$s_2$, /$s_3$ and /p denote decoding from 2nd/3rd semantics ($s_3$) and by probability voting from both semantics respectively, 
CVAIS means modules C+V+A+I+S and so on.}
	\label{tab:ablation}
\end{threeparttable}
\end{table}

\subsection{The Semantic (S) Module}

The semantic module is inserted optionally to play the role of further fusing the two semantic streams.  When inserted, our model is named VST-F (full); otherwise, it is named VST-B (basic).  The two streams are  the secondary semantic features generated from the intersection module and the tertiary (third) semantic features generated from the second vs-align module respectively.

The layer configureation of this module also follows from the design of module V and I.  Fixed positional encoding is added, but there is no domain or segment embedding because the two streams are both semantic. Since the transformer blocks keep the input shape, we concatenate the two streams along channel direction to $t \times 2d $ and feed it to linear and softmax layer, obtaining the final text predictions.

\begin{figure*}[h]
	\includegraphics[width=\textwidth, trim=0.0cm 1.0cm 0.0cm 0.0cm]{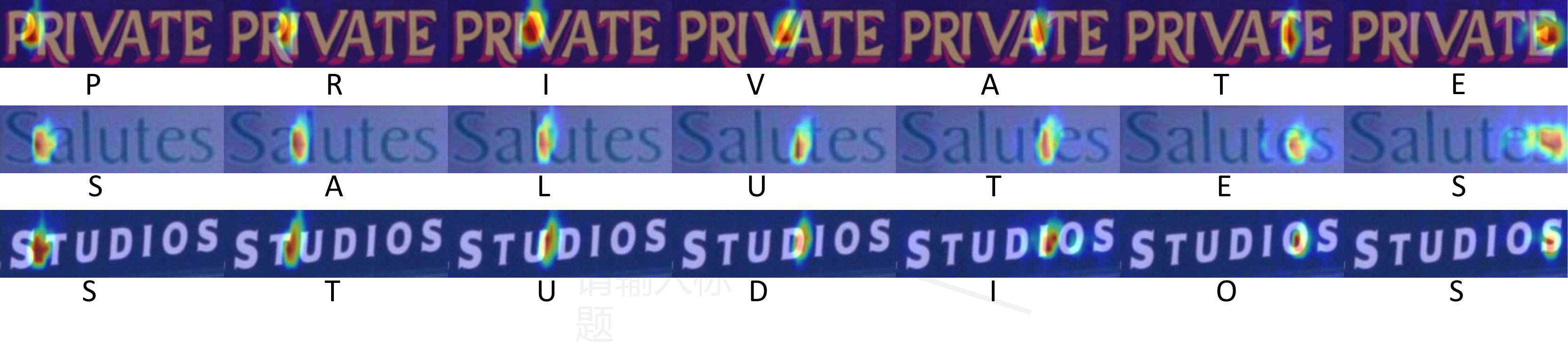}
	\caption{Visualization of attention maps for decoding each character.  }	
	\label{fig:atten_map}
\end{figure*}.

\begin{figure*}[h]
	\includegraphics[width=17.3cm, trim=0.0cm 1.0cm 0.0cm 0.0cm,clip]{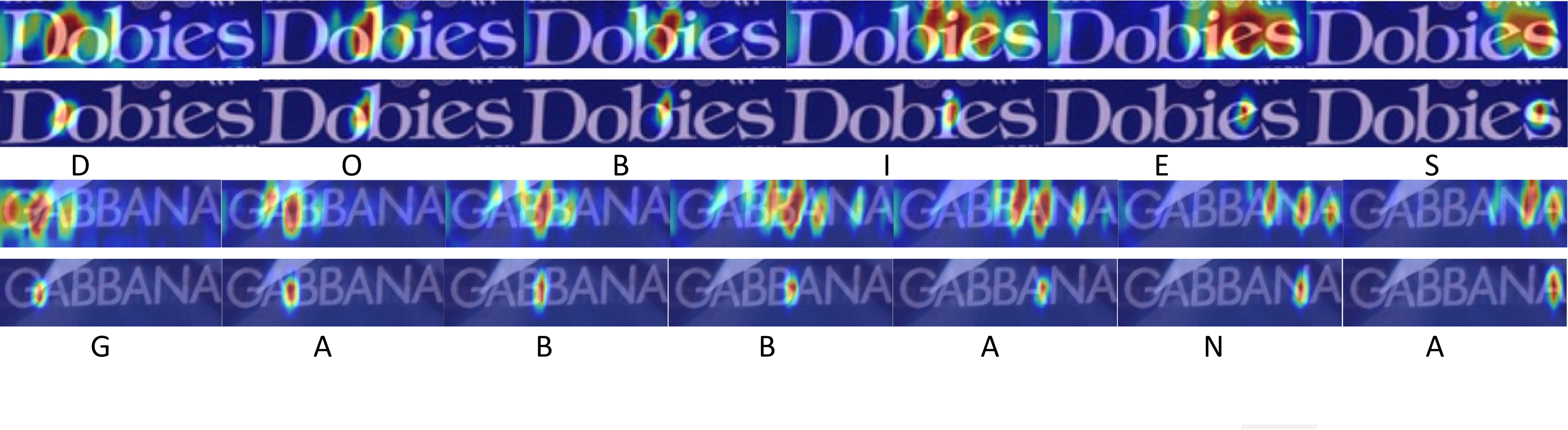}
	\caption{Visualization of attention heatmaps of the primary and secondary vs-align modules. For each of the two examples,
  the top (bottom) row shows the heatmaps of the primary (secondary) vs-align module.  }	
  
	\label{fig:module_i_attn}
\end{figure*}

\subsection{Mathematical interpretation}
As shown in Fig. \ref{arch}, there are three losses indicated by the ladders in different colors.
The green one measures the discrepancy between the  the true text labels and semantic features output from second vs-align module  using softmax activation and cross-entropy loss.  The blue one measures how different are the secondary semantic features from the true text.  The yellow one is the main loss for decoding final text for VST-F,  while the other two are auxiliary for improving convergence performance.  For VST-B, since there is no module S, only blue and green loss are used.  For all loss branches, the discrepancy  are measured by cross-entropy loss with softmax activation.

From the optimization perspective, the VST-F solves  the following optimization problem, 
\begin{equation}
\begin{aligned}
\min_{\theta_v,\theta_a, \theta_i,\theta_s}  \quad & L(s_2,y) + L(s_3,y)+L( f_{\theta_s}(s_2, s_3),y)\\
\textrm{subject to } &\quad v_1 = f_{\theta_v}(x),s_1  = f_{\theta_a}(v_1)\\
& \quad s_2,v_2 = f_{\theta_i}( s_1,v_1), s_3 = f_{\theta_a}(v_2)
\end{aligned}
\end{equation}
where $\theta_v,\theta_a, \theta_i,\theta_s$ denote the parameters of module C+V, A, I, and S respectively, $L$ is cross-entropy loss, $s_1,s_2,s_3,v_1,v_2$ are the primary/secondary/tertiary semantic/visual features resp.,  $x,y$ are the input mage and text label resp. For VST-B, the third term and the corresponding constraint in the above objective function is discarded.

\section{Experiments}
\subsection{Datasets}
Our model is trained on three datasets: SynthText (ST) \cite{ST}, MJSynth (MJ) \cite{MJ-1,MJ-2} and SynthAdd (SA) \cite{SAR}. The training dataset  contains totally 15.7 million synthetic images.
 For evaluation, we use four regular text datasets:  IIIT 5K-words (IIIT) \cite{III-5K} , Street View Text (SVT) \cite{SVT}, ICDAR 2003 (IC03) \cite{IC03}, ICDAR 2013 (IC13) \cite{IC13} ,  and three irregular text datasets:  ICDAR 2015 (IC15) \cite{IC15}, Street View Text Perspective (SVTP) \cite{SVTP}  and CUTE \cite{CUTE} .

IIIT contains $3000$ cropped images collected from Google image search. SVT is collected from Google Street View containing $647$ testing images. Following from \cite{wang2011end}, we select $867$ cropped images from IC03 for testing.  IC13 contains $1095$ testing images and we discard images containing non-alphanumeric characters or contains fewer than three characters, resulting in $1015$ images for testing, following from previous work \cite{SRN}.

For irregular datasets, IC15 contains $2077$ cropped images by Google Glasses. We use $1811$ images after discarding extremely distorted images. SVTP and CUTE contains $639$ and $288$ images respectively. 

\begin{figure}[h]
	\includegraphics[width=0.48\textwidth]{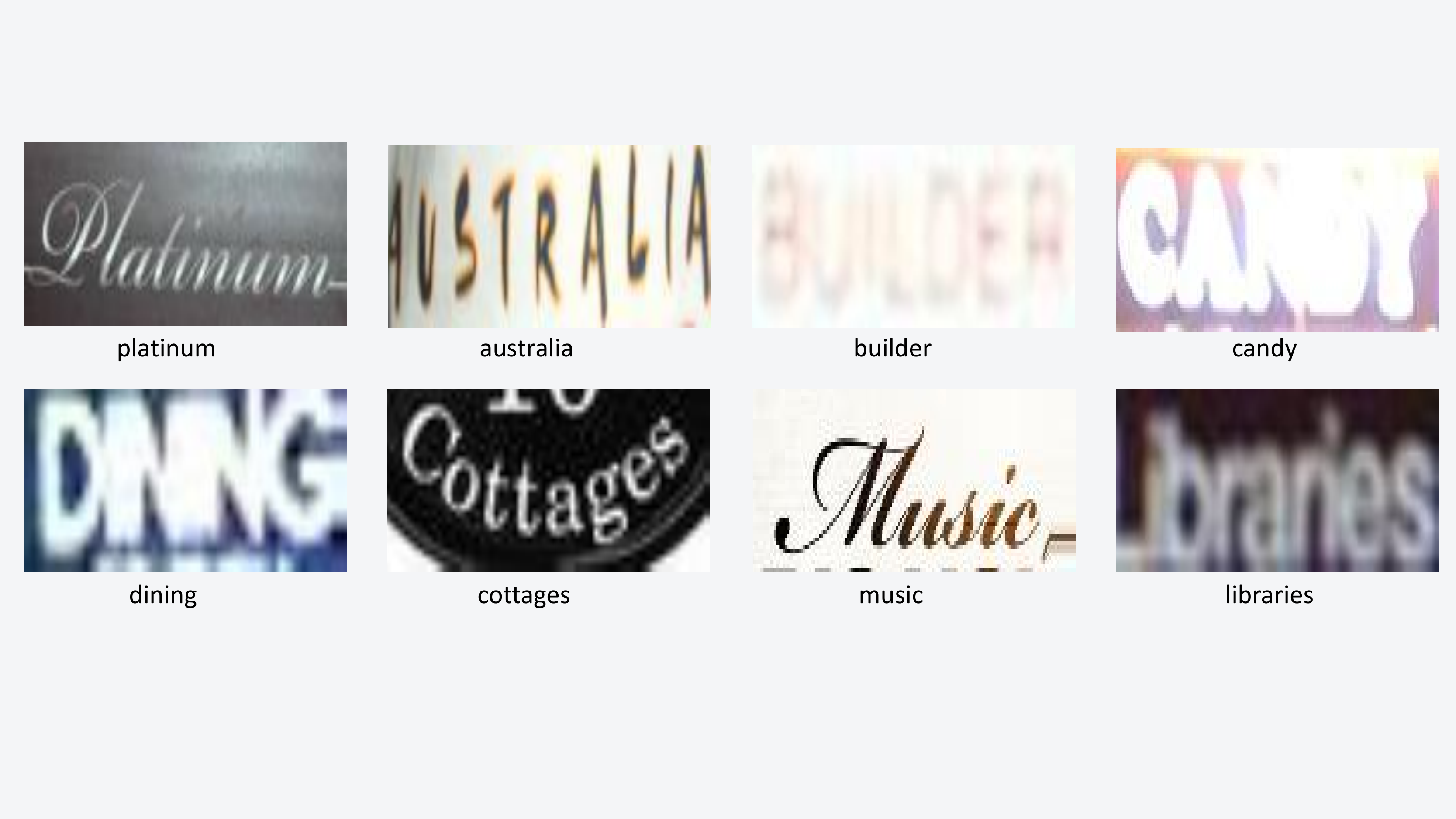}
	\caption{Successful cases for fairly hard examples.}	
	\label{fig:success}
\end{figure}

\begin{figure}[h]
	\includegraphics[width=0.48\textwidth]{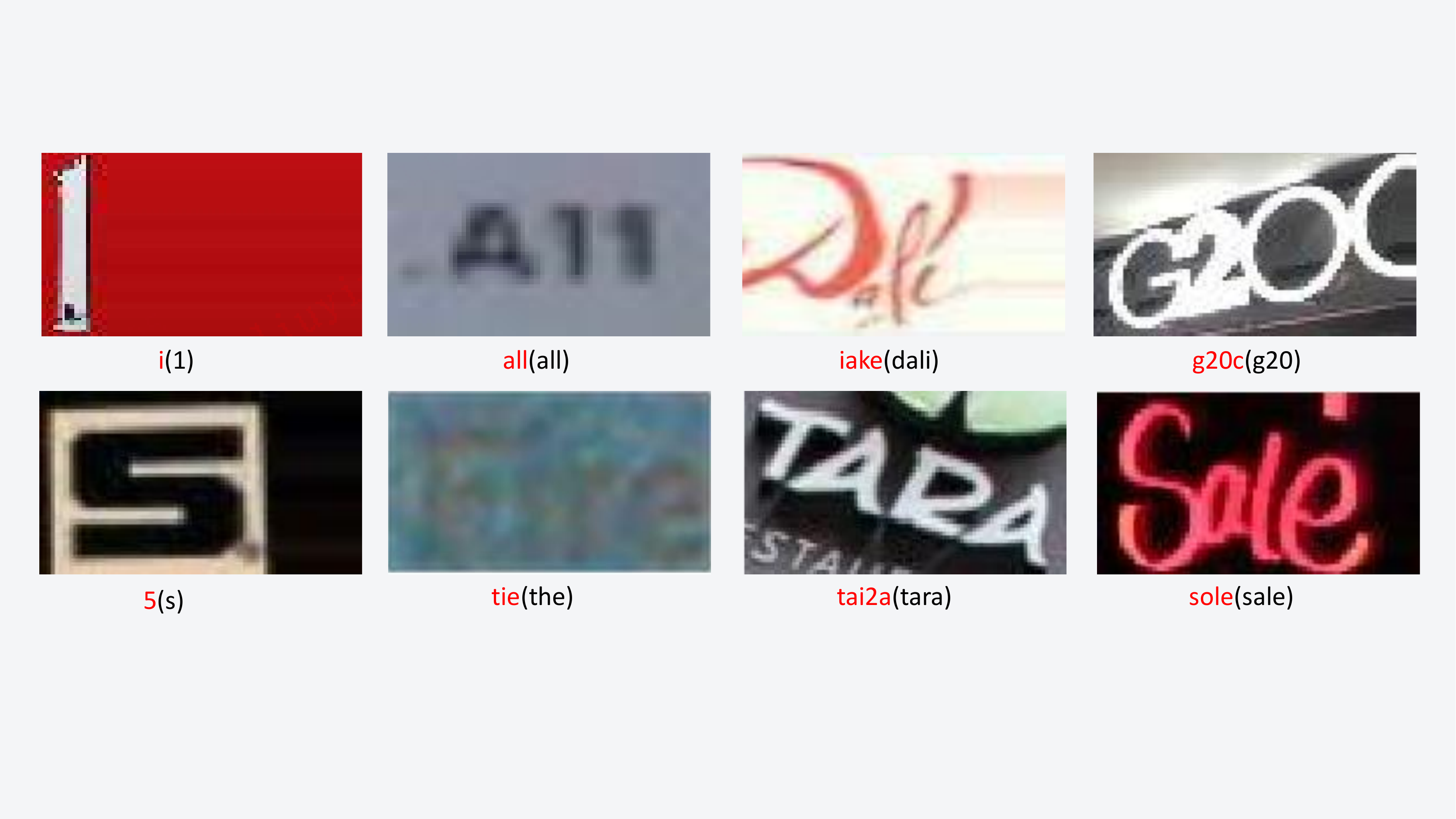}
	\caption{Failure cases for some confusing images.  }	
	\label{fig:failure}
\end{figure}

\subsection{Implementation Details}

Module C is implemented as a four-layers resnet, with  each layer having 1,2,5 and 3 blocks respectively.  Module V, I and S each consists of 3 transformer layers.  The number of parameters are about 63.99M for VST-F, and 63.65M for VST-B.

The image patches are resized with aspect-ratio unchanged so that $H=48$, and the width is trimed or padded to $W=160$ pixels.  The feature map size is  $6 \times 40 \times 512$, $w=40, h=6, n=240$
The number of class is set to 38, including 0-9,a-z,[unk],[eos].  We assume that the maximum character length is 25, i.e., $t=25$. 
We use  online data augmentations including distortion, stretch, perspective transform, blurring, colour jitter etc.  
In training, we set the batch size to 256 and sample  from ST, MJ, SA datasets with weight 0.4,0.4 and 0.2 to balance the datasets \cite{TR}.
We use Adam optimizer with the initial learning rate 1e-4 and decreased lr to 1-e5 when the loss plateaus.  
We use  for  and the training takes roughly 3 days to  converge on four Tesla V100 GPUs.

\subsection{Comparison with State-of-the-Art}
We compare VST-B and VST-F with previous work over seven public benchmarks  in table \ref{tab:sota}. The proposed VST-B and VST-F both achieve  or on par with state-of-the-art approaches. The large version of the very recent work by \cite{RLH}  performs slightly better than ours on three of the seven datasets, but they used extra curbersome pretrained language models. Moreover, our VST-B and VST-F are about three times faster than theirs in inference.  On the rest four datasets, we improve SOTA  by large margins (0.2\% IIIT, 0.3\% SVT, 0.8\% IC03, 2.8\% CUTE).

\subsection{Ablation Study}
To verify whether each module is critical for the final performance, ablation study is conducted by subsequently adding one module starting from the most basic C+V configuration. The results are shown in table  \ref{tab:ablation}, from which we observe a consistent performance gain when a new module is added.  Also note that C+V+A also performs better than C+V, which verifies the effectiveness of vs-align module.  For the C+V+A+I configuration, there are three ways of decoding: decoding from secondary semantics ($s_2$),  tertiary semantics ($s_3$) or by probability voting from both semantics.

\subsection{Visualization}
Inside the module I, each semantic vector will attend to visual features corresponding to some spatial locations. We can visualize which locations are attended to using heapmaps. Fig. \ref{fig:atten_map} shows three examples. The heatmaps are computed by averaging all 8 attention heads and overlapping onto images after  upsampling. Note that inside the module each head focuses on different aspects of the images, but on average the attention is mainly at  the right spatial location for each  decoded character.  For some characters, the focus is slightly shifted, probably due to the translation-invariant property of convnet, which means that the visual features cannot be mapped back uniformly to the corresponding spatial locations. Overall, the visualization  proves that the interaction module works as expected.  

Likewise, we also visualize the attention heatmaps of the primary and secondary vs-align modules. Even though the two modules share the same weights, their attention maps should be different since the visual and semantic features are different. We expect that the attention maps of 2nd vs-align be more precise than the first one.  This is verified in  Fig.~\ref{fig:module_i_attn}.

Beside visualization, we select some successful and failure cases for illustration purpose, shown in Fig. \ref{fig:success} and Fig. \ref{fig:failure} respectively.

\section{Conclusion}
In this paper, we propose visual-semantic transformer to solve scene text recognition problem. The VST consists of  three model explicitly extract semantic features in the early stage by using weight-sharing visual-semantic alignment module, making the visual and semantic fusion and interaction possible int the following transforms modules. 
Our model is unified and end-to-end trainable, and it does not require autoregressive decoding. Extensive experiments on regular and irregular scene text recognition datasets have verified the effectiveness of our model. 
\bibliography{aaai22}

\end{document}